\documentclass[10pt]{article}
\usepackage[letterpaper,margin=0.85in]{geometry}
\newif\ifanonymized
\anonymizedfalse
\newif\ifshortpaper
\shortpaperfalse
\usepackage[T1]{fontenc}
\usepackage[utf8]{inputenc}
\usepackage{newtxtext,newtxmath}
\usepackage{microtype}
\usepackage{amsmath,mathtools}
\usepackage{booktabs}
\usepackage{array}
\usepackage{graphicx}
\usepackage{caption}
\usepackage{subcaption}
\usepackage{enumitem}
\usepackage{url}
\usepackage{xurl}
\usepackage{hyperref}
\usepackage{cleveref}
\usepackage{natbib}
\usepackage{xcolor}
\usepackage{siunitx}
\usepackage{multirow}
\usepackage{tabularx}
\usepackage{makecell}
\usepackage{etoolbox}
\usepackage{ifthen}
\usepackage{balance}
\usepackage{fancyvrb}
\usepackage{placeins,float}

\hypersetup{
  colorlinks=true,
  linkcolor=black,
  citecolor=black,
  urlcolor=blue,
  pdfborder={0 0 0}
}
\setlist[itemize]{leftmargin=*,topsep=3pt,itemsep=2pt,parsep=0pt}
\setlist[enumerate]{leftmargin=*,topsep=3pt,itemsep=2pt,parsep=0pt}
\newcommand{\method}{\textsc{PriorProof}}
\newcommand{\mathlib}{\textsc{Mathlib}}
\newcommand{\lean}{\textsc{Lean}}
\newcommand{\prebin}{\ensuremath{\mathcal{L}_{<b(t)}}}
\newcommand{\footprint}{\ensuremath{\Phi_t(D)}}
\newcommand{\score}{\ensuremath{S_t(D)}}
\newcommand{\gap}{\ensuremath{\Delta(A,B)}}

\newcolumntype{Y}{>{\raggedright\arraybackslash}X}
\newcolumntype{C}[1]{>{\centering\arraybackslash}p{#1}}

\hypersetup{pdftitle={PriorProof: A Point-in-Time Measure of Technique Novelty for Formal Proofs},pdfauthor={Neel Somani},pdfsubject={Formal proof novelty and dependency surprisal},pdfkeywords={Lean, Mathlib, formal mathematics, novelty, surprisal}}

\title{\vspace{-1.5em}\textbf{PriorProof: A Point-in-Time Measure of Technique Novelty for Formal Proofs}}
\author{Neel Somani\\Independent Researcher\\\href{mailto:neeljaysomani@gmail.com}{neeljaysomani@gmail.com}\\Code: \url{https://github.com/neelsomani/priorproof}\\Paper snapshot: \href{https://github.com/neelsomani/priorproof/tree/v1.0-arxiv}{\texttt{v1.0-arxiv}}}
\date{July 2026}

\begin{document}
\maketitle
\begin{abstract}
Mathematicians distinguish proofs that explain, simplify, or introduce a nonstandard route, but these judgments are difficult to operationalize. We study a deliberately narrower construct: \emph{time-relative proof-route nonstandardness} in formal mathematics. For a \lean{} theorem, \method{} extracts the dependency footprint of its elaborated proof term and scores the weighted surprisal of that footprint under a retrieval-conditioned, hierarchically smoothed prior built only from an earlier quarterly snapshot of \mathlib{}. The method requires no hand-built technique ontology and no human labels: statement retrieval is learned from proof-derived contrastive pairs, while the scored object is read mechanically from proof terms. In a blinded topology study, 100 presentations collapse to 76 distinct underlying pairs: 12 canonical contrasts shown three times for consistency screening and 64 distinct stratified pairs. Against the majority of three retained domain raters, \method{} agrees on 53/76 pairs (69.7\%, Wilson 95\% CI 58.7--78.9\%), including 11/12 canonical pairs (91.7\%, 64.6--98.5\%) and 42/64 stratified pairs (65.6\%, 53.4--76.1\%). Score-gap quartiles are nonmonotone after repeat collapse; the endpoints are 12/19 (63.2\%, 41.0--80.9\%) in the smallest-gap bin and 16/19 (84.2\%, 62.4--94.5\%) in the largest, supporting an endpoint-calibration tendency rather than a resolved staircase. The best language-model condition agrees on 60/76 pairs (78.9\%, 68.5--86.6\%); on paired outcomes, \method{} alone is correct on 8 pairs and the model alone on 15 (exact two-sided McNemar $p=0.210$), so the difference is not established at this sample size. We therefore present \method{} not as a replacement for expert or model judgment, but as a decomposable, time-anchored signal whose score gap provides an interpretable reliability indicator.

\end{abstract}
\section{Introduction}
\label{sec:intro}

Formal proof libraries expose a record that informal mathematics usually does not: an elaborated term identifies the constants used by a proof, and version control identifies which declarations existed beforehand. This makes formal mathematics a useful controlled setting for a question that is otherwise hard to pose mechanically: how nonstandard was a proof route relative to the techniques available when the proof was written?

The question concerns one proof virtue among several. Philosophers and mathematicians have separately analyzed explanation, purity of method, and elegance \citep{steiner1978,lange2014,detlefsen2011,inglis2015}. Route novelty is not reducible to any of them. A routine statement can admit an unexpected proof, while an important theorem can be established through familiar machinery. The judgment is also temporal: a route that is surprising when introduced may later become standard.

This distinction is increasingly relevant as learned theorem provers produce formal proofs at scale \citep{lample2022hypertree,yang2023leandojo,xin2024deepseekprover,lamont2025threeD}. Existing evaluations primarily ask whether a theorem is proved, how much search is required, or how well a model transfers. They do not directly ask whether a successful proof follows a route that the earlier library would have made likely. A mechanical signal for that question could support analysis of machine-generated proofs, historical studies of formal libraries, and high-recall triage for expert review.

\begin{figure}[t]
  \centering
  \includegraphics[width=\linewidth]{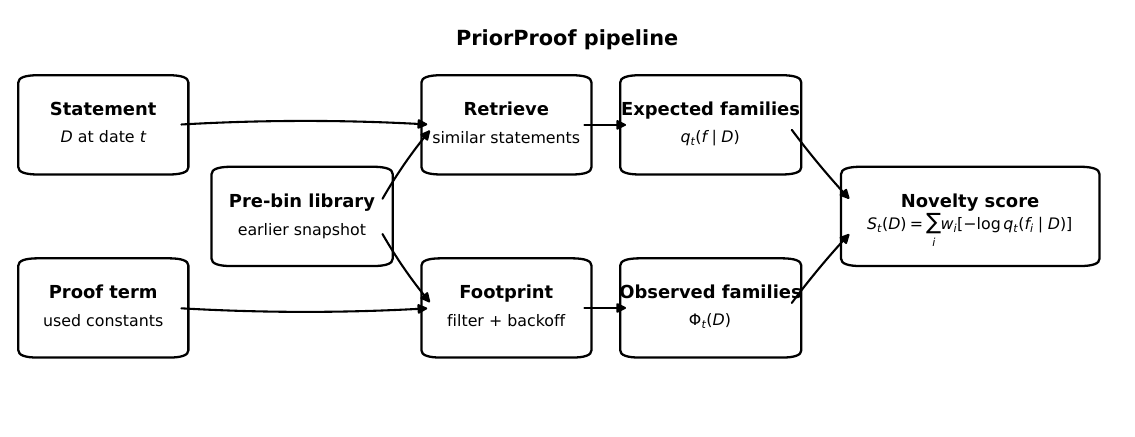}
  \caption{\method{} separates what was \emph{expected} from what was \emph{used}. Statement-only retrieval over a pre-bin library induces a prior $q_t(f\mid D)$ over dependency families, while the elaborated proof term yields an observed footprint $\footprint$. Their weighted surprisal is the score.}
  \label{fig:pipeline}
\end{figure}

We introduce \method{}, a quarterly point-in-time measure of \emph{proof-route nonstandardness} for \lean{}~4 / \mathlib{} \citep{demoura2021lean4,mathlib2020}. For a declaration $D$ written at time $t$, the pipeline (\cref{fig:pipeline}) performs three operations:
\begin{enumerate}
  \item extract a weighted dependency-family footprint from the elaborated proof term;
  \item retrieve similar theorem statements from a library snapshot strictly earlier than the start of $D$'s quarter and construct a smoothed prior over dependency families; and
  \item score the weighted surprisal of the observed footprint under that prior.
\end{enumerate}
A high score means that the proof uses families of prior machinery that were improbable for earlier theorems with similar statements. The score is about the proof \emph{as written}; it is not a claim about the author's originality, the theorem's importance, or the proof's explanatory value.

Our contributions are:
\begin{itemize}
  \item a mechanical, annotation-free definition of time-relative proof-route nonstandardness as dependency-footprint surprisal;
  \item a temporal leakage discipline in which retrieval, family support, reuse counts, and encoder fine-tuning are restricted to pre-bin data;
  \item an empirical study on \mathlib{} topology, including mechanical validation, a blinded three-rater comparison, and an identically prompted language-model baseline; and
  \item evidence for an \emph{endpoint-calibration tendency}: the smallest score-gap bin is statistically compatible with chance and the largest has the highest agreement point estimate, but intervals are wide at this sample size.
\end{itemize}

The headline result should be read narrowly. Overall agreement with the three-rater majority is 53/76 (69.7\%, Wilson 95\% CI 58.7--78.9\%), and the broad stratified subset is 42/64 (65.6\%, 53.4--76.1\%). The canonical behavior check is 11/12 (91.7\%, 64.6--98.5\%), but its distinct sample is small and its interval wide. The strongest conclusion is not universal ranking accuracy but a candidate reliability indicator in this sample: the absolute score gap. To our knowledge, prior work has not scored an existing formal proof by the surprisal of its dependency footprint under a statement-conditioned prior restricted to the library's earlier state. This is an operationalization of one component of technique novelty, not a complete theory of mathematical novelty.

\section{Related work}
\label{sec:related}

\ifshortpaper
\else
\paragraph{Premise selection and retrieval.}
Learning-assisted premise selection ranks prior facts that may help prove a target theorem, from hammer-style systems for Isabelle/HOL to neural retrieval in large formal corpora \citep{blanchette2016,irving2016deepmath,bansal2019holist}. LeanDojo combines retrieval with language-model proof search in \lean{} \citep{yang2023leandojo}, while miniCTX studies proving with evolving, long-form context and releases \texttt{ntp-toolkit} for extracting \lean{} data \citep{hu2024minictx}. \method{} uses retrieval for a different estimand: not to propose a proof, but to define what dependency families an earlier library would have made expected for a statement.

\paragraph{Learned theorem proving.}
Neural provers have advanced through learned proof search, retrieval augmentation, synthetic data, and diversity-aware search \citep{lample2022hypertree,yang2023leandojo,xin2024deepseekprover,lamont2025threeD}. These systems motivate the evaluation question addressed here: conditional on success, did a model recover a standard route or use unexpectedly different machinery? \method{} is model-agnostic after a proof has been checked by \lean{}.

\paragraph{Formal-library structure and proof dependencies.}
Lean and \mathlib{} provide the elaborated objects and library organization on which our extractor operates \citep{demoura2021lean4,mathlib2020}. LeanDojo and miniCTX demonstrate large-scale extraction and temporally aware theorem-proving datasets \citep{yang2023leandojo,hu2024minictx}. Work on proof structures and the network organization of \mathlib{} studies dependency graphs as mathematical objects \citep{wernhard2024proofstructures,li2026networkmathlib}; concurrent work such as TheoremGraph uses graph representations to connect formal and informal mathematics \citep{kurgan2026theoremgraph}. Our target differs: a declaration-level, time-sliced surprisal score for the route taken by an existing proof.

\paragraph{Novelty and surprisal.}
Surprisal is the negative log probability of an observation \citep{shannon1948}. Related operationalizations appear in anomaly detection \citep{ruff2021} and bibliometrics, where atypical combinations of references are used as a signal of scientific novelty \citep{uzzi2013}. The analogy is useful but incomplete. A formal proof supplies mechanically checked dependencies rather than author-selected citations, and a versioned library permits a stricter temporal counterfactual.

\paragraph{Proof virtues.}
Work on mathematical explanation, purity, and aesthetic appraisal emphasizes that proof quality is multidimensional \citep{steiner1978,lange2014,detlefsen2011,inglis2015}. We do not collapse these virtues into one score. \method{} targets only whether the proof route uses unexpectedly distributed prior machinery.
\fi

\section{Method}
\label{sec:method}

\subsection{Temporal object and notation}

Let $D$ be a theorem declaration with statement $x_D$, elaborated proof term $p_D$, and commit time $t$. Let $b(t)$ be the beginning of the quarterly snapshot bin containing $t$, and let
\begin{equation}
  \prebin = \{E : \operatorname{time}(E) < b(t)\}
  \label{eq:prebin}
\end{equation}
be the library available to the implementation. Conceptually, the desired counterfactual is the library immediately before $t$; the implementation uses the stricter pre-bin slice in \cref{eq:prebin}. This prevents target leakage but excludes declarations introduced earlier in the same quarter, so temporal resolution is quarterly rather than commit-exact.

\subsection{From a proof term to a dependency footprint}
\label{sec:footprint}

Surface tactics such as \texttt{simp} or \texttt{linarith} are poor units of technique: they are syntactic interfaces whose mathematical work depends on the lemmas and instances eventually elaborated. At the opposite extreme, raw constant names are too specific and over-credit proofs that create fresh helper objects. We therefore construct a weighted family footprint in four stages.

\paragraph{Actual-use extraction.}
For each compiled theorem value, the proof-term backend calls \lean{}'s \texttt{getUsedConstants}. This yields constants actually referenced by the declaration's elaborated value, rather than all imports of its source file. In the reported run, every normalized dependency reference is marked as proof-term-derived (\cref{sec:setup}).

\paragraph{Deterministic filtering.}
A documented filter removes generated recursors, coercions, routine typeclass and notation plumbing, and common tactic boilerplate. This is a design choice rather than a theorem about mathematical relevance; filter variants are therefore part of the validation interface.

\paragraph{Established-machinery frontier.}
A proof-local or recently introduced dependency is recursively replaced by the older dependencies on which it rests until the traversal reaches declarations established in \prebin. Establishment is defined by a pre-bin reuse-count threshold. Reuse is never counted in a snapshot containing the target, preventing a batch of co-introduced declarations from making one another appear established. Surviving dependencies receive deterministic inverse-frequency-style weights from the implementation.

\paragraph{Family backoff.}
Each surviving constant maps to the finest family with at least five pre-bin occurrences, backing off through declaration, parent namespaces, module, library area, and global levels. This reduces accidental novelty from rare names without introducing a separately annotated technique taxonomy. The resulting footprint is
\begin{equation}
  \footprint = \{(f_i,w_i)\}_{i=1}^{m_D}, \qquad w_i > 0,
  \label{eq:footprint}
\end{equation}
where $f_i$ is a supported dependency family and $w_i$ is its deterministic footprint weight.

\subsection{Statement encoder and pre-bin retrieval}
\label{sec:encoder}

The retriever embeds only the theorem statement $x_D$, never $p_D$. It begins from \nolinkurl{sentence-transformers/all-MiniLM-L6-v2}, a compact sentence-embedding model descended from MiniLM \citep{reimers2019sbert,wang2020minilm}. Contrastive fine-tuning examples are mined mechanically from the pre-bin formal corpus. Positive pairs share signals such as dependency families, downstream users, major dependency links, or namespace-local symbols. Hard negatives include lexically similar statements with disjoint families, cross-module false friends, and statements with similar heads but different shapes. Training uses Multiple Negatives Ranking Loss for one epoch with batch size 64, AdamW at learning rate $2\times10^{-5}$, 10\% linear warmup, a 256-token maximum sequence length, shuffled batches, and hard negatives enabled. The device is selected automatically by Sentence-Transformers. No training seed was set, so the learned embedding is nondeterministic across runs.

Because the training signal is proof-derived, an encoder trained on future proofs could leak future dependency structure into retrieval. The pipeline therefore mines pairs and trains a separate encoder for each scoreable bin using only declarations before that bin. A cheaper shared-encoder path is permitted by the code only after a neighbor-stability test; in the reported run, cross-bin overlap was 0.43 against a prespecified 0.75 threshold, so per-bin encoders were enforced.

For target $D$, the encoder retrieves $k=32$ statement-nearest theorems from \prebin{} by cosine similarity of $\ell_2$-normalized embeddings. Neighbor evidence is weighted by a softmax with fixed temperature 0.2. Their proofs contribute empirical mass to the families they used. Retrieval is thus conditioned on the target statement while remaining independent of the target proof.

\subsection{Hierarchically smoothed prior}
\label{sec:prior}

Let $q_{r,t}$, $q_{n,t}$, $q_{m,t}$, and $q_{g,t}$ denote empirical family distributions derived respectively from retrieved neighbors, the target namespace, the target module, and the full pre-bin library. The prior has the mixture form
\begin{equation}
 q_t(f\mid D)=\lambda_r q_{r,t}(f\mid D)+\lambda_n q_{n,t}(f\mid D)
   +\lambda_m q_{m,t}(f\mid D)+\lambda_g q_{g,t}(f),
 \label{eq:prior}
\end{equation}
with nonnegative weights summing to one and additive smoothing parameter $\alpha>0$. The components are computed from pre-bin weighted family counts. Mixture weights and $\alpha$ are selected by chronological log likelihood: hide each eligible proof, build its prior only from earlier declarations, and maximize the probability assigned to the families the proof actually used. This fitting objective is predictive rather than tuned to the human study.

\subsection{Novelty score and pairwise confidence signal}

The \method{} score is the realized weighted surprisal of the footprint:
\begin{equation}
  \score = \sum_{i=1}^{m_D} w_i\big[-\log q_t(f_i\mid D)\big].
  \label{eq:score}
\end{equation}
This is not a cross-entropy against a separate reference distribution; it is the weighted surprisal of the observed families. Every contribution can be decomposed into a named dependency family, its weight, and its pre-bin probability.

For a pair $(A,B)$, the metric selects the proof with larger score. Its internal confidence signal is the absolute score gap
\begin{equation}
  \gap = |S_{t_A}(A)-S_{t_B}(B)|.
  \label{eq:gap}
\end{equation}
The human study tests whether larger gaps correspond to more reliable pairwise judgments.

\subsection{Leakage discipline}
\label{sec:leakage}

The pre-bin slice mechanically excludes the target, its formal descendants, future retrieval candidates, future reuse counts, and future family-support evidence. Two residual channels require separate treatment.

First, the pretrained encoder initialization can contain parametric knowledge of later mathematics. This cannot be removed by slicing. The counterfactual-retrieval probe replaces relevant retrieved context with unrelated pre-bin context; a score that changes demonstrates dependence on retrieved evidence, whereas invariance flags cases in which retrieval is not materially influencing the output. The probe measures sensitivity but does not prove the absence of parametric memory.

Second, proof-derived encoder fine-tuning can transmit future dependency patterns. Per-bin training from \prebin{} addresses this channel mechanically. A remaining unobservable channel is diffuse human influence: a later proof may use an idea learned from $D$ without a dependency edge to $D$. Such influence would make an earlier route appear more standard than it was, biasing the metric downward rather than manufacturing novelty.

\subsection{Directness assumption and redundancy backstop}
\label{sec:directness}

\method{} scores the proof as written. It is meaningful when the proof is reasonably direct, rather than padded with an eliminable detour that re-derives an available theorem through exotic machinery. The pipeline implements two redundancy checks against \prebin: a top-level path for wrappers that reduce to a single prior theorem, and a nested path that compares sub-derivation conclusions with prior statements up to trivial equivalence. Both paths pass a constructed re-derivation fixture. The live proof-term backend, however, emits top-level dependency sets but no proof subterms, so the nested path is dormant in the reported corpus. This backstop catches redundancy with the existing library, not arbitrary baroqueness around genuinely new intermediates.

\subsection{Scope of the construct}
\label{sec:scope}

The score omits several phenomena often called novelty. It is blind to novelty residing primarily in definitions; it measures unexpected dependency families rather than unexpected \emph{connections} between otherwise familiar families; and it does not target elegance, explanatory value, difficulty, importance, or authorship credit. Point-in-time scoring also means later simplification does not retroactively change the score assigned at introduction.

\section{Experimental setup}
\label{sec:setup}

\subsection{Corpus and provenance}

The study uses target declarations under \path{Mathlib.Topology.MetricSpace}, \path{Mathlib.Topology.Compactness}, and \path{Mathlib.Topology.Separation}, with neighboring topology modules retained as support context. \Cref{tab:corpus} summarizes the corpus. The 2023Q4 snapshot is the first conditioning baseline; targets in 2024Q1--2024Q2 are scoreable because they have an earlier in-manifest corpus. Each snapshot is pinned to the latest commit on the \mathlib{} \texttt{master} branch at or before the quarter start: 2023Q4 on 2023-10-01, 2024Q1 on 2024-01-01, and 2024Q2 on 2024-04-01. Full commit hashes and \lean{} toolchains are reported in \cref{app:provenance}.

\begin{table}[H]
  \caption{Corpus and extraction summary. Support declarations condition retrieval and smoothing but are not part of the reported target population.}
  \label{tab:corpus}
  \centering
  \small
  \begin{tabular}{lr}
    \toprule
    Quantity & Count \\
    \midrule
    Scoped declarations & 10,132 \\
    Target declarations & 2,026 \\
    Support declarations & 8,106 \\
    Scored post-baseline targets & 248 \\
    Normalized proof-term dependency references, full extraction & 4,707,281 \\
    Proof-term dependency references, topology scope & 264,914 \\
    Records with extracted proof subterms & 0 \\
    \bottomrule
  \end{tabular}
\end{table}

The extraction report records \texttt{backend: proof-term}; all 4,707,281 normalized dependency references are tagged \texttt{source: proof\_term}. The absence of proof subterms explains why nested redundancy detection is fixture-validated but inactive on live data.

\subsection{Mechanical validation}

We evaluate six properties before comparing with raters: (i) chronological likelihood of observed versus randomly chosen families; (ii) retrieval and smoothing ablations; (iii) counterfactual-retrieval sensitivity; (iv) score sensitivity to footprint construction; (v) redundancy-fixture behavior; and (vi) correlation between score and family-backoff depth. The reuse threshold is swept over $\{3,5,8,13\}$ with a bucket-identity diagnostic. Because every declaration maps to identical final family buckets in this sweep, it is reported as inert rather than as independent robustness evidence.

\subsection{Blinded human study}
\label{sec:human}

The evaluation packet contains 100 blinded \emph{presentations} of 76 distinct underlying proof pairs. Twelve hand-selected canonical contrasts are each shown three times, giving 36 presentations, and 64 score-gap-stratified corpus pairs are each shown once. The canonical repeats were included deliberately as a consistency instrument rather than as independent outcome replicates. For each side, raters see the declaration name, theorem statement, proof narrative, and complete \lean{} source. They do not see scores, family probabilities, or canonical/stratified labels. The prompt is:
\begin{quote}
\emph{Which proof uses the less standard mathematical route to its result?}
\end{quote}
The wording deliberately avoids asking which proof is more ``original.''

Outcome analysis is performed on the 76-pair universe. Each left/right choice is first normalized to the underlying proof identity, and each distinct pair is assigned a canonical binary orientation. For each judge, the primary label for a repeated canonical pair is the majority choice across its three presentations; the 64 singleton pairs pass through unchanged. No collapse ties occur for any retained rater or language-model condition. A first-presentation-only sensitivity analysis changes every reported aggregate by at most one pair. \method{} is deterministic across presentations by construction.

Four topology mathematicians completed the packet. One response set was screened post hoc after the sequence described in \cref{sec:screening}; all headline outcome statistics use the three retained raters. The screening response is reported separately and never enters the majority labels used below.

The task was conducted by an independent researcher with no institutional affiliation requiring institutional review. It involved expert judgments over public mathematical content, posed minimal risk, and collected no sensitive research data. Raters were told that their responses would be used in research. Names and payment details were collected only for recruitment and payment, retained privately, and omitted from every released artifact. Each completed packet was compensated at \$1,500; the estimated completion time was roughly two hours (about \$750/hour, and \$500/hour under a three-hour completion assumption). Every recruited rater was paid in full, including the screened rater. Human and model responses remain private because the consent language did not address public release.

\subsection{Language-model baseline}
\label{sec:llm}

Judgments were collected through the OpenAI Responses API under the provider model names \texttt{gpt-5} and \texttt{gpt-5-mini}, crossed with two prompt-strictness variants, producing 400 judgments over the same 100 blinded presentations. No date-pinned model identifiers were available in the run records, and temperature was not explicitly set, so the API default applied. For each condition, repeated presentations are collapsed by the same proof-identity and majority procedure used for human judges. A four-condition majority vote is then formed on the 76 distinct pairs, excluding 2--2 ties. Requests and responses for the designated run and a second identical-setting run are retained privately; the two runs of the best logged condition disagreed on 17 of the 100 presentations. Execution dates were not logged, so the baseline should be interpreted as a contemporaneous provider-name comparison rather than an exactly reproducible model benchmark. Proof narratives shown to both human and model judges were generated with \texttt{gpt-5-mini} under a system instruction that prohibited formal-system metadata, evaluation vocabulary, and judgments about whether a route was standard or surprising; outputs were checked against forbidden terms and patterns and retried on failure. We cite the model-family documentation for context \citep{openai2025gpt5}, not as evidence for the reported judgments.

\subsection{Statistics and availability}

Agreement proportions and Wilson score intervals are computed on distinct underlying pairs after repeat collapse \citep{wilson1927}. Pairwise rater agreement is reported with Cohen's $\kappa$ \citep{cohen1960}; three-rater agreement uses Fleiss' $\kappa$ \citep{fleiss1971}. All $\kappa$ calculations use a canonical binary orientation defined separately for each proof pair. The comparison between \method{} and the best language-model condition uses an exact two-sided McNemar test on the same 76 distinct pairs. Wilson intervals remain descriptive because all pairs share a small rater panel and the canonical subset is hand-selected, but repeated presentations are not treated as independent outcomes.

\ifanonymized
\else
Code and the blinded study-packet machinery are available under the MIT License at \url{https://github.com/neelsomani/priorproof}. The results reported in this paper correspond to Git tag \href{https://github.com/neelsomani/priorproof/tree/v1.0-arxiv}{\texttt{v1.0-arxiv}}. Human and language-model response data are not public because the rater disclosure did not cover public release; consequently, the released pipeline reproduces the method and can generate a fresh study through the same tooling, but it does not reproduce the exact reported judgment aggregates.
\fi

\section{Results}
\label{sec:results}

\subsection{Mechanical validation}

\ifshortpaper
\else
Observed footprint families receive higher chronological probability than random alternatives: mean log probability $-10.132$ versus $-12.056$, a margin of 1.924 nats over 5,629 footprint items. The topology-specific fit is strongly retrieval-weighted, whereas a broader all-\mathlib{} fit is more balanced (\cref{tab:prior}). This difference cautions against treating the topology mixture as universal.

\begin{table}[H]
  \caption{Chronologically fitted prior mixtures. The topology slice relies heavily on retrieved neighbors; the broader fit assigns more mass to structural backoff.}
  \label{tab:prior}
  \centering
  \small
  \begin{tabular}{lccccc}
    \toprule
    Fit & Retrieval & Namespace & Module & Global & $\alpha$ \\
    \midrule
    Topology & 0.90 & 0.04 & 0.03 & 0.03 & 0.05 \\
    All \mathlib{} & 0.40 & 0.25 & 0.20 & 0.15 & 0.10 \\
    \bottomrule
  \end{tabular}
\end{table}

Ablations raise observed-footprint surprisal on average, but effects are uneven across declarations (\cref{tab:ablations}). The no-namespace ablation is especially weak: its mean is positive but only 40\% of declarations move upward, consistent with namespace smoothing acting mostly as a small floor in this slice. The counterfactual-retrieval probe has no empty-retrieval rows among the 248 scored targets; mean signed sensitivity is 1.31 and mean absolute sensitivity is 1.59. Backoff depth is essentially uncorrelated with score ($r=0.046$, $n=992$ family observations). There are no live redundancy hits. All 10,132 scoped declarations have identical family buckets across reuse thresholds 3, 5, 8, and 13.

\begin{table}[H]
  \caption{Ablations relative to the fitted topology prior. Positive $\Delta$ means greater surprisal on observed footprints after removing a component.}
  \label{tab:ablations}
  \centering
  \small
  \begin{tabular}{lrr}
    \toprule
    Ablation & Mean $\Delta$ surprisal & Fraction with $\Delta>0$ \\
    \midrule
    Global only & +5.49 & 0.65 \\
    No retrieval & +1.85 & 0.52 \\
    No module & +0.71 & 0.51 \\
    No namespace & +0.34 & 0.40 \\
    \bottomrule
  \end{tabular}
\end{table}
\fi

\subsection{Rater screening and repeat consistency}
\label{sec:screening}

The quality screen was developed post hoc. Outcome statistics were initially computed with all four raters. One response set was then flagged by a median response time of roughly seven seconds per presentation for a packet designed around 90--120 seconds of reading, by weak agreement with the other human and automated judges, and by the rater's subsequent explanation. The order-normalized repeat-consistency statistic was constructed last and became the criterion of record.

For each triplicated canonical pair, choices are normalized to proof identity. The within-judge majority identifies the reference choice for that group. The vote score counts how many of the 36 presentations match that majority, so its mechanical floor is 24/36; a random binary responder has expectation 27/36. The group score counts fully consistent triplicates, with random-response expectation 3/12. \Cref{tab:consistency} reports the result.

\begin{table}[H]
  \caption{Order-normalized repeat consistency. Vote counts are matches to each judge's within-group majority; group counts are fully consistent triplicates. The language-model row gives the range over all four model--prompt conditions.}
  \label{tab:consistency}
  \centering
  \small
  \begin{tabularx}{\linewidth}{Ycc}
    \toprule
    Judge & Majority-match votes /36 & Fully consistent groups /12 \\
    \midrule
    \method{} (by construction; order-invariant) & 36 & 12 \\
    Retained rater A & 35 & 11 \\
    Retained rater B & 35 & 11 \\
    Retained rater C & 33 & 9 \\
    Four language-model conditions (range) & 33--35 & 9--11 \\
    Screened rater & 28 & 4 \\
    Random binary response baseline & floor 24; $\mathbb{E}=27$ & $\mathbb{E}=3$ \\
    \bottomrule
  \end{tabularx}
\end{table}

The screened response yields 28/36 majority-match votes and 4/12 fully consistent groups. Under random binary responding, $\Pr(X\geq4)=0.351$ for the group count, so this statistic alone does not strongly reject chance and lies within roughly one standard deviation of its expectation. The retained raters yield 35, 35, and 33 votes and 11, 11, and 9 fully consistent groups; even the lowest retained group count has chance probability approximately $4\times10^{-4}$. The screen cannot distinguish a random responder from a fixed position-biased responder, but either pattern indicates failure to engage with proof identity; the seven-second median supplies independent engagement evidence. All headline results use the three retained raters. The language-model conditions yield 33--35 votes and 9--11 groups, so the language-model baseline survives the study's own consistency instrument and cannot be dismissed on the same ground.

\subsection{Rater agreement}

On the 76 distinct pairs, the three retained rater-pair Cohen coefficients are $0.548$, $0.461$, and $0.550$. Fleiss' $\kappa$ is $0.519$ overall, $0.750$ on the 12 canonical pairs, and $0.476$ on the 64 stratified pairs. All three raters are unanimous on 49/76 pairs (64.5\%, Wilson 95\% CI 53.3--74.3\%). Thus, the broad stratified subset is a moderately noisy construct target rather than ground truth. The $\kappa$ calculation uses a per-pair binary orientation; treating proof identities as global categories would make chance agreement artificially tiny and inflate $\kappa$ toward raw agreement.

\subsection{Agreement with the rater majority}
\label{sec:majority}

\Cref{tab:accuracy} reports agreement with the three-rater majority after canonical repeats are collapsed. \method{} selects the majority-preferred route on 53/76 distinct pairs (69.7\%, Wilson 95\% CI 58.7--78.9\%). Agreement is 11/12 on the canonical behavior check (91.7\%, 64.6--98.5\%) and 42/64 on the broader stratified sample (65.6\%, 53.4--76.1\%). The canonical point estimate is high, but its interval is wide because the distinct canonical set contains only 12 pairs.

\begin{table}[H]
  \caption{Agreement with the three-rater majority on distinct underlying pairs after repeat collapse. The canonical set is a behavior check; the stratified set is the broader generalization estimate.}
  \label{tab:accuracy}
  \centering
  \small
  \begin{tabular}{lrrl}
    \toprule
    Subset & Correct & Agreement & Wilson 95\% CI \\
    \midrule
    All distinct pairs & 53/76 & 69.7\% & [58.7\%, 78.9\%] \\
    Canonical & 11/12 & 91.7\% & [64.6\%, 98.5\%] \\
    Stratified & 42/64 & 65.6\% & [53.4\%, 76.1\%] \\
    \bottomrule
  \end{tabular}
\end{table}

\subsection{Endpoint-calibration tendency by score gap}
\label{sec:calibration}

\begin{figure}[H]
  \centering
  \includegraphics[width=0.80\linewidth]{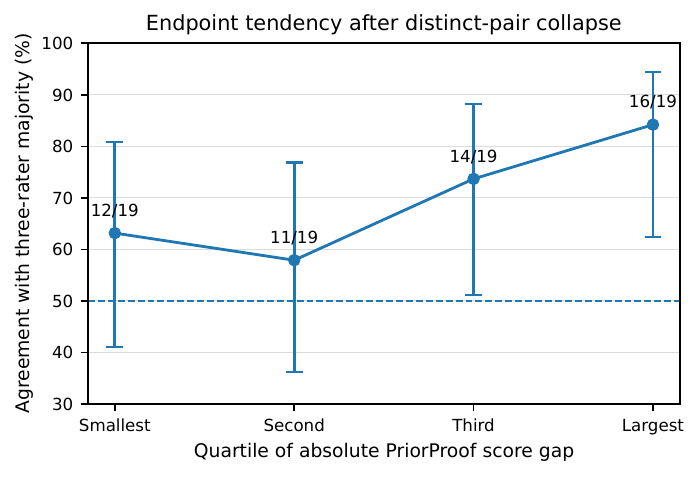}
  \caption{Agreement with the three-rater majority by quartile of absolute score gap on 76 distinct pairs ($n=19$ per bin). Error bars are Wilson 95\% intervals. After repeated canonical presentations are collapsed, the interior bins are nonmonotone; the evidence is an endpoint tendency, not a resolved staircase.}
  \label{fig:calibration}
\end{figure}

After collapse, the 76 distinct pairs divide evenly into four score-gap bins of 19. Agreement is 12/19 in the smallest-gap bin (63.2\%, Wilson 95\% CI 41.0--80.9\%), 11/19 in the second (57.9\%, 36.3--76.9\%), 14/19 in the third (73.7\%, 51.2--88.2\%), and 16/19 in the largest (84.2\%, 62.4--94.5\%). The interior point estimates are nonmonotone, and all intervals are wide. The evidence therefore supports only an endpoint tendency: the smallest-bin interval includes chance, while the largest bin has the highest point estimate and excludes 50\%. High-gap comparisons are the more reliable regime in this sample, but no resolved four-step calibration staircase is established.

\subsection{Language-model comparison}
\label{sec:llmresults}

The best logged language-model condition (\texttt{gpt-5-mini}, strict prompt) agrees with the rater majority on 60/76 distinct pairs (78.9\%, Wilson 95\% CI 68.5--86.6\%). The four-condition majority, excluding five 2--2 ties, agrees on 57/71 non-tied pairs (80.3\%, 69.6--87.9\%). Both point estimates are nominally higher than \method{}'s 53/76 result (69.7\%, 58.7--78.9\%).

On the 76 paired outcomes for the best condition, \method{} alone is correct on 8 pairs and the language-model judge alone is correct on 15. The exact two-sided McNemar test gives $p=0.210$; the discordant counts and $p$-value are identical under the primary majority collapse and the first-presentation-only sensitivity rule. The paired test therefore does not resolve the difference at this sample size, but its direction is clear: the language-model judge has the higher nominal pairwise-agreement estimate. As \cref{tab:consistency} shows, every language-model condition also survives the repeat-consistency screen. Across two identical-setting API runs, the best condition changes its decision on 17 of 100 presentations, a presentation-level nondeterminism diagnostic rather than a distinct-pair accuracy result. \method{}'s value consequently rests on properties the model judgment lacks: a fixed point-in-time corpus, mechanical decomposition, and an explicit score-gap signal.

\section{Discussion}
\label{sec:discussion}

\ifshortpaper
\else
\paragraph{What the calibration result buys.}
A scalar ranking metric is most useful when it indicates when its own ordering is fragile. In this study, the largest score-gap bin has the highest agreement point estimate and excludes chance, while the smallest-gap bin does not; the two interior bins are nonmonotone and the intervals are wide. This supports a selective workflow in which high-gap contrasts receive priority for automatic surfacing and low-gap cases are deferred to domain judgment. It does not establish a universal threshold, and a second library area is required before treating the pattern as transferable.

\paragraph{Interpretability.}
A score decomposes into family-level terms $w_i[-\log q_t(f_i\mid D)]$. This makes it possible to inspect whether a high score is driven by a genuinely unexpected mathematical area, a coarse backoff family, or an implementation artifact. The backoff-depth diagnostic reduces one obvious confound, but interpretability does not make the family hierarchy uniquely correct.

\paragraph{Relation to learned provers.}
The natural downstream experiment is to score multiple checked proofs of the same theorem: human proofs, model-generated proofs, and search variants. Because the score is evaluated after proof checking, it can compare provers without changing their search procedure. Care is essential, however: machine-generated proofs may exploit detours or library idiosyncrasies more aggressively than curated \mathlib{} proofs, making normalization and nested redundancy detection prerequisites for strong claims.

\paragraph{Why this is not an originality detector.}
Unexpected dependencies can arise from genuine invention, an awkward proof, a deliberate obfuscation, missing library lemmas, or a mismatch between the target and retrieval corpus. Conversely, an original conceptual insight can elaborate to standard dependencies. The measure is therefore evidence about a formal route, not evidence about plagiarism, priority, intent, or deservingness of credit.
\fi

\section{Limitations}
\label{sec:limitations}

\ifshortpaper
\else
\paragraph{Construct validity.}
Dependency-family surprisal is only one component of technique novelty. The metric misses definitional novelty and connection-level novelty, and it does not measure explanation, elegance, difficulty, or importance. The family hierarchy inherits \mathlib{}'s namespaces and modules, which are organizational rather than a canonical ontology of mathematical technique.

\paragraph{Temporal precision and leakage.}
The implementation is quarterly. Using the start of the target's bin is leakage-safe but may overstate novelty by omitting legitimate earlier declarations from the same quarter. Pretrained encoder memory is measured only indirectly through context sensitivity, not eliminated. Diffuse human-mediated influence without a formal edge remains unobservable and biases novelty downward.

\paragraph{Proof normalization.}
The score applies to proofs as written. The live extractor exposes no subterms, leaving nested redundancy detection inactive. A contrived or machine-generated proof can inflate its score by importing unusual machinery or taking a long detour. The zero live-hit result is evidence about the detector's visible target class in this curated corpus, not proof that every proof is direct.

\paragraph{Statistical scope.}
The evaluation covers 248 targets in one \mathlib{} domain, three retained mathematicians, and 76 distinct comparison pairs. Only 12 canonical contrasts are distinct; their three presentations are a consistency instrument and must not be treated as separate outcomes. Counting repeats as observations would overstate both precision and apparent performance. Agreement is weakest on the 64-pair stratified subset. The screening rule was constructed after inspection, and the screened rater's 4/12 fully consistent groups have $\Pr(X\geq4)=0.351$ under random binary responding; the exclusion therefore also depends on the approximately seven-second median and the broader engagement evidence. The score-gap result is an endpoint tendency with wide intervals, not a resolved calibration curve.

The language-model judge has the higher nominal point estimate, survives the same repeat-consistency instrument, and has 15 paired-only wins versus 8 for \method{}; the exact McNemar test does not resolve the difference at this sample size. The contribution should therefore be judged by the metric's mechanical provenance, temporal anchoring, decomposition, and candidate reliability signal, not by a claim to match or beat the model judge. A second library area and a substantially larger distinct canonical set, with repeats retained as a consistency instrument, are structural requirements before any cross-domain claim.

\paragraph{Baseline and reproducibility metadata.}
The \mathlib{} commit hashes, \lean{} toolchains, encoder hyperparameters, retrieval settings, prior grid, and MIT code license are recorded. Exact date-pinned language-model identifiers and execution timestamps were not logged; encoder package versions, checkpoint hashes, pair-sampling seeds, and measured hardware/runtime logs were also not retained. Human and model responses are private, so the released code cannot regenerate the exact reported judgment tables, although item-level records retained by the author support controlled reanalysis. These gaps limit exact bitwise reproduction and external audit but do not alter the supplied aggregates.

\paragraph{Adversarial use.}
The metric can be gamed by adding irrelevant rare machinery. Conversely, coarse backoff can suppress a subtle but meaningful novelty. It should not be optimized as a standalone reward until proof normalization and gaming tests are substantially stronger.
\fi

\section{Ethics, intended use, and broader impact}
\label{sec:ethics}

\method{} is intended as a research instrument for analyzing checked formal proofs and prioritizing cases for expert inspection. Appropriate uses include historical library analysis, comparing proof-search systems, and selecting high-gap examples for qualitative study. Inappropriate uses include automated plagiarism findings, authorship attribution, hiring or promotion decisions, or allocation of mathematical credit. Such decisions require evidence about provenance, intent, and conceptual contribution that the score does not contain.

The system processes public formal-library artifacts. The expert-annotation study was conducted by an independent researcher outside an institution requiring review, was minimal risk, and collected no sensitive research data. Raters were informed that their judgments would be used in research and were compensated \$1,500 per completed packet; all four were paid in full. Names and payment details remain private and no identifying information appears in released artifacts. Individual responses are withheld because public release was not covered by the rater disclosure. This protects privacy but limits auditability; a future study should obtain consent for de-identified release or establish a controlled-access protocol.

The main foreseeable negative impact is false authority: a quantitative ``novelty'' label may be over-interpreted. We mitigate this by naming the narrower construct, reporting broad-pair performance and rater disagreement, exposing the weaker endpoint-calibration tendency, and explicitly prohibiting person-level conclusions. A positive impact is more disciplined evaluation of machine-generated mathematics. Rather than treating proof success as the sole outcome, researchers could examine whether successful systems merely reproduce expected dependency patterns or discover routes that merit human attention. The metric should serve as a hypothesis generator, not an arbiter.

\ifanonymized\else
\paragraph{Funding and competing interests.} This work received no external funding and was self-funded by the author, including rater compensation. The author declares no competing interests.

\paragraph{AI-assistance disclosure.} Portions of the manuscript prose and analysis scripts were drafted with LLM assistance; all claims, numbers, and citations were verified by the author against primary artifacts and sources.
\fi

\section{Conclusion}
\label{sec:conclusion}

We presented \method{}, a mechanical signal of quarterly point-in-time proof-route nonstandardness. The method compares a proof-term dependency footprint with a statement-conditioned prior induced solely from an earlier \mathlib{} snapshot. On topology, it agrees with a three-rater majority on 53/76 distinct pairs (69.7\%, Wilson 95\% CI 58.7--78.9\%), including 11/12 canonical contrasts (91.7\%, 64.6--98.5\%) and 42/64 stratified pairs (65.6\%, 53.4--76.1\%). The largest score-gap bin has the highest point estimate, 16/19 (84.2\%, 62.4--94.5\%), versus 12/19 (63.2\%, 41.0--80.9\%) in the smallest; the nonmonotone interior bins and wide intervals make this an endpoint tendency rather than a resolved calibration curve. The language-model judge has a higher nominal estimate, and an exact paired test does not resolve the difference at this sample size. The result supports cautious selective use of the metric when score gaps are large, not unqualified ranking of arbitrary proofs. Future work should add finer-grained temporal slicing, subterm-aware proof normalization, connection-level priors, a second mathematical domain, a larger distinct canonical set, preregistered rater screening, and a consented or controlled-access route for paired human/model judgments.

\bibliographystyle{plainnat}
\bibliography{references}
\clearpage
\appendix

\section{Algorithmic specification}
\label{app:algorithm}

For a target declaration $D$ in bin $b(t)$, the released pipeline implements the following high-level procedure:
\begin{enumerate}
  \item Check out the library snapshot at the beginning of the bin and restrict every corpus-derived quantity to \prebin.
  \item Extract actual used constants from $p_D$ with the proof-term backend.
  \item Apply the deterministic plumbing filter.
  \item Recursively unfold non-established dependencies to the established-machinery frontier using pre-bin reuse counts.
  \item Map frontier constants to supported families by hierarchical backoff and assign deterministic footprint weights, yielding \cref{eq:footprint}.
  \item Embed $x_D$ with the encoder trained only on data before the bin and retrieve nearest statements from \prebin.
  \item Build the mixture prior in \cref{eq:prior} from retrieved, namespace, module, and global family counts.
  \item Compute \cref{eq:score}; emit the total and per-family contributions.
\end{enumerate}

The implementation additionally records the snapshot ID, encoder path, family-backoff level, retrieval neighborhood, prior mixture, redundancy flags, and counterfactual score, enabling audit of each score.

\section{Experimental provenance}
\label{app:provenance}

\subsection{Pinned \mathlib{} snapshots and \lean{} toolchains}

Each snapshot uses the latest commit from the \mathlib{} \texttt{master} branch at or before the bin start; extraction was performed on the listed start date. The toolchain is the value in the pinned commit's \texttt{lean-toolchain} file.

\begin{table}[htbp]
  \caption{Snapshot dates and \lean{} toolchains.}
  \centering
  \small
  \begin{tabular}{lll}
    \toprule
    Snapshot & Start / extraction date & \lean{} toolchain \\
    \midrule
    2023Q4 & 2023-10-01 & \texttt{leanprover/lean4:v4.2.0-rc1} \\
    2024Q1 & 2024-01-01 & \texttt{leanprover/lean4:v4.5.0-rc1} \\
    2024Q2 & 2024-04-01 & \texttt{leanprover/lean4:v4.7.0-rc2} \\
    \bottomrule
  \end{tabular}
\end{table}

\noindent Full \mathlib{} commit identifiers:
\begin{description}[style=nextline,leftmargin=1.2cm,labelwidth=1.0cm]
  \item[2023Q4] \texttt{aef04106feb057e57456331886e5f38e392dea9f}
  \item[2024Q1] \texttt{2a17457d3236d97eec6687377c01a74fe2961ab7}
  \item[2024Q2] \texttt{d8d7e696a4c05914b5f2dbff8768541fe1dd4b39}
\end{description}

\subsection{Encoder training}

\begin{table}[htbp]
  \caption{Per-bin statement-encoder configuration. No seed was set.}
  \centering
  \small
  \begin{tabularx}{\linewidth}{p{0.34\linewidth}Y}
    \toprule
    Field & Value \\
    \midrule
    Base model & \nolinkurl{sentence-transformers/all-MiniLM-L6-v2} \\
    Objective & Multiple Negatives Ranking Loss, with mined hard negatives \\
    Training slice & Contrastive examples restricted to declarations before the score bin \\
    Epochs / batch size & 1 / 64 \\
    Optimizer & AdamW (Sentence-Transformers default, with learning rate overridden) \\
    Learning rate / warmup & $2\times10^{-5}$ / 10\% linear warmup \\
    Maximum sequence length & 256 tokens \\
    Data order & Shuffled \\
    Device & Auto-selected by Sentence-Transformers (CUDA when available, otherwise CPU) \\
    Seed & None; DataLoader order and accelerator execution may vary across runs \\
    \bottomrule
  \end{tabularx}
\end{table}

Exact Sentence-Transformers/PyTorch versions, accelerator model, and checkpoint hashes were not retained in the supplied experimental record. The disclosed configuration therefore supports methodological reproduction but not bitwise reconstruction of the original learned embeddings.

\subsection{Retrieval, family backoff, and prior grid}

Retrieval uses $k=32$, cosine similarity over $\ell_2$-normalized embeddings, and softmax temperature 0.2. A raw declaration family is used only after at least five historical occurrences; otherwise the implementation backs off through parent namespaces, module, area, and global levels. The prior grid crosses
\[
  \alpha\in\{0.01,0.025,0.05,0.1,0.25,0.5\}
\]
with the mixture candidates in \cref{tab:priorgrid}. Parameters are selected by chronological log likelihood, independently of the human judgments.

\begin{table}[htbp]
  \caption{Candidate prior mixtures $(\lambda_r,\lambda_n,\lambda_m,\lambda_g)$.}
  \label{tab:priorgrid}
  \centering
  \small
  \begin{tabular}{lcccc}
    \toprule
    Candidate & Retrieval & Namespace & Module & Global \\
    \midrule
    Default & 0.55 & 0.20 & 0.15 & 0.10 \\
    Balanced & 0.40 & 0.25 & 0.20 & 0.15 \\
    Retrieval-heavy I & 0.70 & 0.10 & 0.10 & 0.10 \\
    Retrieval-heavy II & 0.80 & 0.08 & 0.06 & 0.06 \\
    Retrieval-heavy III & 0.90 & 0.04 & 0.03 & 0.03 \\
    Retrieval only & 1.00 & 0.00 & 0.00 & 0.00 \\
    No retrieval & 0.00 & 0.35 & 0.30 & 0.35 \\
    \bottomrule
  \end{tabular}
\end{table}

The selected topology configuration is $(0.90,0.04,0.03,0.03)$ with $\alpha=0.05$; the selected all-\mathlib{} configuration is $(0.40,0.25,0.20,0.15)$ with $\alpha=0.10$. Reuse thresholds $\{3,5,8,13\}$ were swept, but family assignments were identical throughout the reported corpus.

\subsection{Compute and randomness record}

Wall-clock logs were not retained. The author records the encoder fine-tuning as a single-consumer-GPU workload on the order of minutes per bin and extraction as a CPU workload on the order of hours per snapshot. Hardware model, memory, storage, exact stage times, and total exploratory compute are unknown. Nondeterministic components are: unseeded encoder fine-tuning; language-model API judgments (17 of 100 presentation-level decisions changed across two identical-setting runs of the best condition); and potentially packet sampling, for which a seed was not recovered from the supplied records. The committed packet artifact fixes the evaluated sample even where the generating seed is unavailable.

\section{Mechanical-validation details}
\label{app:mechanical}

\ifshortpaper
\else
The fitted prior mixtures and ablations are reported in \cref{tab:prior,tab:ablations}; they are not duplicated here. The remaining mechanical diagnostics are: observed-family mean log probability $-10.132$ versus $-12.056$ for random alternatives over 5,629 footprint items; counterfactual-retrieval signed and absolute sensitivities 1.31 and 1.59 over 248 targets; score/backoff-depth correlation $r=0.046$ over 992 family observations; zero live redundancy hits; and identical family buckets for all 10,132 declarations across reuse thresholds 3, 5, 8, and 13.
\fi

\FloatBarrier
\section{Complete result tables}
\label{app:results}

All outcome summaries in this section use the 76 distinct underlying pairs after the 12 triplicated canonical groups are collapsed. The consistency instrument itself is reported separately in \cref{tab:consistency}.

For the best language-model condition, the paired discordances are 8 \method{}-only correct pairs and 15 model-only correct pairs; the exact two-sided McNemar test gives $p=0.210$. The primary majority collapse and first-presentation-only sensitivity rule produce the same discordant counts and $p$-value. Across all other aggregate outcomes, the first-presentation sensitivity differs from the primary collapse by no more than one pair.

\begin{table}[!htbp]
  \caption{Score-gap results used in \cref{fig:calibration}. Each bin contains 19 distinct pairs.}
  \centering
  \small
  \begin{tabular}{lrrl}
    \toprule
    Gap quartile & Correct & Agreement & Wilson 95\% CI \\
    \midrule
    Smallest & 12/19 & 63.2\% & [41.0\%, 80.9\%] \\
    Second & 11/19 & 57.9\% & [36.3\%, 76.9\%] \\
    Third & 14/19 & 73.7\% & [51.2\%, 88.2\%] \\
    Largest & 16/19 & 84.2\% & [62.4\%, 94.5\%] \\
    \bottomrule
  \end{tabular}
\end{table}

\begin{table}[!htbp]
  \caption{Retained-rater agreement on distinct pairs. Cohen's $\kappa$ values are computed under a canonical binary orientation for each pair.}
  \centering
  \small
  \begin{tabular}{lll}
    \toprule
    Quantity & Estimate & Wilson 95\% CI, if applicable \\
    \midrule
    Cohen's $\kappa$, three rater pairs & 0.548; 0.461; 0.550 & --- \\
    Fleiss' $\kappa$, all distinct pairs & 0.519 & --- \\
    Fleiss' $\kappa$, canonical & 0.750 & --- \\
    Fleiss' $\kappa$, stratified & 0.476 & --- \\
    Three-rater unanimity & 49/76 = 64.5\% & [53.3\%, 74.3\%] \\
    \bottomrule
  \end{tabular}
\end{table}

\begin{table}[!htbp]
  \caption{Language-model comparison on distinct pairs. Five 2--2 ties are excluded from the four-condition majority row.}
  \centering
  \small
  \begin{tabular}{lrrl}
    \toprule
    Condition & Correct & Agreement & Wilson 95\% CI \\
    \midrule
    Best logged condition: \texttt{gpt-5-mini}, strict & 60/76 & 78.9\% & [68.5\%, 86.6\%] \\
    Four-condition majority, non-tied pairs & 57/71 & 80.3\% & [69.6\%, 87.9\%] \\
    \method{} & 53/76 & 69.7\% & [58.7\%, 78.9\%] \\
    \bottomrule
  \end{tabular}
\end{table}

\FloatBarrier
\section{Human-study packet, screening, and ethics}
\label{app:study}

Each comparison side contains four fields: declaration name, theorem statement, a natural-language proof narrative, and full \lean{} source. The public rater packet excludes metric scores, priors, family buckets, metric-derived explanations, and the canonical/stratified source label. Narratives are generated from the blinded packet rather than from score artifacts. The same materials and core question are passed to human and language-model judges.

The packet contains 100 presentations of 76 distinct underlying pairs. Twelve canonical pairs are each presented three times; 10 of the 12 groups include a left/right swap and two retain the same order. The remaining 64 stratified pairs are unique. Repeats were deliberately included for consistency measurement. For outcome analysis, left/right choices are normalized to proof identity, each pair receives a canonical binary orientation, and each judge's repeated group is collapsed by majority. No collapse ties occur for retained raters or model conditions, and the 64 singleton pairs remain unchanged.

The quality screen was developed post hoc. Outcome statistics were first computed with all four raters. One response set was then flagged by approximately seven-second median response time, weak agreement with other human and automated judges, and an explanation solicited from the rater. The order-normalized majority-match statistic was constructed last and used as the criterion of record. The screened rater records 28/36 majority-match votes and 4/12 fully consistent groups, compared with 35/36, 35/36, and 33/36 votes and 11/12, 11/12, and 9/12 groups for the retained raters. Random binary responding has vote floor 24/36, expected votes 27/36, and expected fully consistent groups 3/12; $\Pr(X\geq4)=0.351$. A fixed position bias and random responding are not separable under this instrument, but both fail to track proof identity. The timing evidence is therefore part of the exclusion rationale. All headline outcomes use the three retained raters. The four language-model conditions achieve 33--35 votes and 9--11 groups, so they pass the same consistency instrument.

The study was conducted by an independent researcher with no institutional affiliation requiring review. It involved expert annotation of public formal proofs, was minimal risk, and collected no sensitive research data. Raters were informed that responses would be used in research. Names and payment information were used only for recruitment/payment, retained privately, and omitted from released artifacts. Compensation was \$1,500 per completed packet, and every recruited rater was paid in full. The estimated completion time was roughly two hours; this corresponds to approximately \$750/hour, or \$500/hour under a three-hour completion assumption. Human responses are not released because the disclosure to raters did not address public data release.

Future studies should preregister the exact exclusion rule, preserve item-level timestamps and responses in a dated manifest, enlarge the distinct canonical set while retaining repeats as an instrument, and add another mathematical domain. A hierarchical model with item and rater random effects would better reflect the shared-panel design.

\section{Language-model baseline provenance}
\label{app:llmprovenance}

Judgments were collected through the OpenAI Responses API under the provider names \texttt{gpt-5} and \texttt{gpt-5-mini}; date-pinned model IDs were not available in the run records. Two prompt-strictness levels were crossed with the two models. Temperature was not explicitly set and therefore followed the API default. Each condition's repeated presentations are normalized to proof identity and collapsed before outcomes are scored. A majority over the four conditions excludes 2--2 splits. Request and response files for the designated run and a second identical-setting run are retained privately. The best condition differs on 17 of 100 presentations across those runs.

Proof narratives were generated by \texttt{gpt-5-mini}. The narrative system instruction forbade formal-system metadata, evaluation vocabulary, and any judgment that a proof was standard, surprising, or preferred. Outputs were checked against forbidden-term and forbidden-pattern lists and retried upon failure. Exact prompt strings and pair templates are retained in the request archive but are not reproduced in this source package; they should be deposited verbatim with any venue supplement that requires request-level reproducibility. Execution dates were not logged, and exact-model reproduction cannot be guaranteed because provider names may later resolve to updated snapshots.

\section{Release, licensing, and reproduction map}
\label{app:reproduction}

The PriorProof code is released under the MIT License. The public repository documents five reproduction stages:
\begin{enumerate}
  \item \textbf{Extraction and snapshots:} freeze \mathlib{} commits, run the \lean{} proof-term extractor, normalize declarations, and build pre-bin reuse counts.
  \item \textbf{Footprints:} filter dependencies, unfold to the established frontier, back off to supported families, and write one footprint artifact per reuse threshold.
  \item \textbf{Encoder and prior:} mine pre-bin contrastive pairs, fine-tune per-bin encoders, retrieve statement neighbors, and fit mixture weights by chronological likelihood.
  \item \textbf{Scoring and mechanical validation:} produce scores and priors, run ablations, create counterfactual priors, and aggregate threshold/backoff/redundancy diagnostics.
  \item \textbf{Judgment study:} assemble canonical and stratified presentations, render the blinded rater interface, generate identically structured model requests, and analyze responses after repeat collapse.
\end{enumerate}

The analysis command fails closed when it detects duplicate underlying pairs, including side-swapped duplicates, unless \texttt{--allow-repeats-with-collapse} is supplied. It accepts the screened response separately through \texttt{--screening-response}, asserts \method{}'s determinism across presentations, and verifies that all collapsed labels use the canonical binary orientation. These checks prevent repeated presentations or screened responses from silently entering outcome statistics.

Generated corpora, model checkpoints, answer keys, human responses, and language-model responses are excluded from the small public release. Independent groups can regenerate the method and run a fresh blinded study, but cannot reproduce the exact reported human/model aggregates from public files alone. The source package includes a third-party asset ledger. Exact installed Python package versions, encoder checkpoint hashes, hardware logs, and unlogged API execution dates remain unavailable.

\clearpage
\section{Claim-to-evidence map}
\label{app:claims}

\begin{table}[H]
  \caption{Scope of the paper's principal claims.}
  \centering
  \small
  \begin{tabularx}{\linewidth}{p{0.28\linewidth}Y Y}
    \toprule
    Claim & Evidence & Non-claim \\
    \midrule
    The prior predicts used families better than random alternatives & Chronological mean log-probability margin over 5,629 footprint items & The prior is a calibrated generative model of all valid proofs \\
    Clear contrasts behave as intended & 11/12 = 91.7\% agreement, Wilson 95\% CI [64.6\%, 98.5\%] & The 12-pair canonical estimate represents broad-corpus accuracy \\
    Larger gaps are the more reliable endpoint in this sample & 16/19 = 84.2\% [62.4\%, 94.5\%] in the largest bin versus 12/19 = 63.2\% [41.0\%, 80.9\%] in the smallest & A monotone calibration curve or a transferable threshold \\
    The metric is broadly informative but noisy & 42/64 = 65.6\% [53.4\%, 76.1\%] on stratified pairs; Fleiss' $\kappa=0.476$ & Rater majority is objective ground truth \\
    The model judge has the higher nominal point estimate & 60/76 = 78.9\% [68.5\%, 86.6\%] versus 53/76 = 69.7\% [58.7\%, 78.9\%]; McNemar discordances 8 versus 15, $p=0.210$ & Statistically established superiority, equality, or non-inferiority \\
    Repeats function as a consistency instrument & Retained raters and model conditions have 9--11 fully consistent groups; screened response has 4; random expectation is 3 & A preregistered screen or evidence that the model baseline fails engagement \\
    \bottomrule
  \end{tabularx}
\end{table}

\end{document}